\documentclass[letterpaper]{article} 
\usepackage{aaai25}  
\usepackage{times}  
\usepackage{helvet}  
\usepackage{courier}  
\usepackage[hyphens]{url}  
\usepackage{graphicx} 
\usepackage{natbib}  
\usepackage{caption} 
\usepackage{algorithm}
\usepackage{algorithmic}
\usepackage{amsmath}
\usepackage{amsfonts}
\usepackage{array}       
\usepackage{booktabs}
\usepackage{multirow}
\usepackage{cases}
\usepackage{newfloat}
\usepackage{enumitem}
\usepackage{listings}
\DeclareCaptionStyle{ruled}{labelfont=normalfont,labelsep=colon,strut=off} 
\floatstyle{ruled}
\newfloat{listing}{tb}{lst}{}
\floatname{listing}{Listing}
\nocopyright 

\title{BitQ: Tailoring Block Floating Point Precision for Improved DNN \\ Efficiency on Resource-Constrained Devices}
\author {
    Yongqi Xu\textsuperscript{\rm 1},
    Yujian Lee\textsuperscript{\rm 2},
    Gao Yi\textsuperscript{\rm 1},
    Bosheng Liu\textsuperscript{\rm 1},
    Yucong Chen\textsuperscript{\rm 3},
    Peng Liu\textsuperscript{\rm 1},
    Jigang Wu\textsuperscript{\rm 1},\\
    Xiaoming Chen\textsuperscript{\rm 4},
    Yinhe Han\textsuperscript{\rm 4}
}
\affiliations {
    \textsuperscript{\rm 1}Computer Science and Technology, Guangdong University of Technology\\
    \textsuperscript{\rm 2}Faculty of Science and Technology, Beijing Normal University - Hong Kong Baptist University United International College\\
    \textsuperscript{\rm 3}	School of Information Science and Engineering, Lanzhou University\\
    \textsuperscript{\rm 4}		Institute of Computing Technology, Chinese Academy of Sciences\\
}

\usepackage{bibentry}

\begin{document}

\maketitle

\begin{abstract}
Deep neural networks (DNNs) are powerful for cognitive tasks such as image classification, object detection, and scene segmentation. One drawback however is the significant high computational complexity and memory consumption, which makes them unfeasible to run real-time on embedded platforms because of the limited hardware resources. Block floating point (BFP) quantization is one of the representative compression approaches for reducing the memory and computational burden owing to their capability to effectively capture the broad data distribution of DNN models. Unfortunately, prior works on BFP-based quantization empirically choose the block size and the precision that preserve accuracy. In this paper, we develop a BFP-based bitwidth-aware analytical modeling framework (called ``BitQ'') for the best BFP implementation of DNN inference on embedded platforms. We formulate and resolve an optimization problem to identify the optimal BFP block size and bitwidth distribution by the trade-off of both accuracy and performance loss. Experimental results show that compared with an equal bitwidth setting, the BFP DNNs with optimized bitwidth allocation provide efficient computation, preserving accuracy on famous benchmarks. The source code and data are available at https://github.com/Cheliosoops/BitQ.
\end{abstract}

%

\section{Introduction}
Recent advancements in deep neural networks (DNNs) have captured substantial interest from both academia and industry \cite{cheng2024advancements,liu2024sora,magictime}.
The convergence of vast amounts of data and powerful computational prowess has fueled the rise of DNNs as a dominant force in modern computing \cite{hanna2024does,gao2023opendmc,bassi2024improving}. Nevertheless, one shortcoming of these DNNs pertains to the computational complexity and memory overhead. The expense associated with computation and memory access poses significant challenges when implementing state-of-the-art DNNs in practice on embedded platforms such as autonomous vehicles and mobile devices because of the limited hardware resources \cite{zheng2023research,liu2024dynamic,clements2024resource}. To address such challenge, among the array of compression techniques aimed at enhancing model efficiency, block floating point (BFP) quantization \cite{drumond2018training,kosson2024multiplication,akkad2023embedded} emerges as a prominent strategy for reducing both the memory footprint and computational burden. The BFP-based numerical representation has recently been established and deployed by the industry for the upcoming AI infrastructure \cite{chronomagic}.

BFP quantization provides a compromise between floating-point and fixed-point formats by grouping multiple floating point numbers into a block and replacing the data format with a shared exponent while retaining their respective mantissa \cite{darvish2023shared,song2023refloat,lo2023bucket}. The advantages of BFP quantization are in twofold: it enables the streamlined storage of scale factors per block instead of for each data point individually, and simplifies multiplication by applying scaling factors in blocks rather than individually. BFP quantization combines the advantages of floating-point precision with fixed-point efficiency, providing an appealing solution for applications that prioritize memory efficiency, precision, dynamic range, and computational speed. 


To increase the effective arithmetic density and memory bandwidth, researchers develop low bitwidth BFP quantization to reduce the precision of both weights and activations \cite{low}. 
Different from only focusing on the mantissa component, some works reduce the bitwidth of both the exponent and mantissa components of BFP data representation \cite{lian2019high, drumond2018training}. For example, Drumond et al. \cite{drumond2018training} develop hybrid BFP data representation to maximize fixed-point arithmetic and minimize the mantissa width requirements while preserving convergence. 
 Lian et al. \cite{lian2019high} implement 8-bit BFP quantization with a 5-bit block exponent for DNN deployments, achieving less than 1\% accuracy loss. Different from prior works that only focus on bitwidth quantization, Nascimento et al. \cite{nascimento2023hyperblock}  analyze the accuracy and efficiency impact of block size on BFP quantization. To fully utilize the bitwidth BFP quantization, DBPS \cite{lee2023dbps} involves a dynamic block size and precision scaling to minimize energy consumption. FAST \cite{zhang2022fast} adaptively selects the optimal precision for weights and activations in the calculation. 

\begin{figure}[!t]
\centering
\includegraphics[scale=0.72]{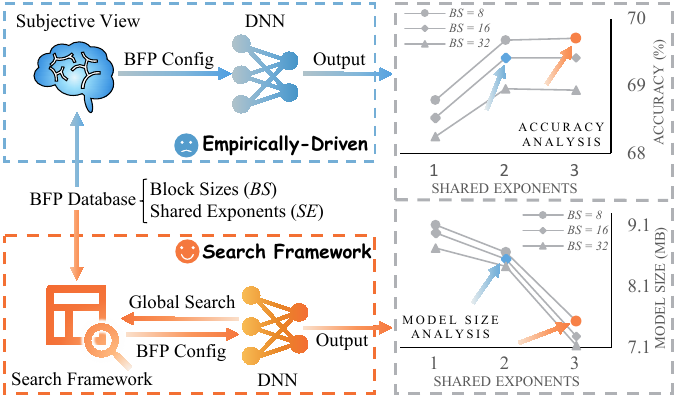}
\caption{Comparison of BFP quantization configuration selection strategies: The figure illustrates the empirically-driven strategy in the upper section and the search framework strategy in the lower section, with line charts showcasing the performance gap on ResNet-18 \cite{resnet18} between the two selection methods.}
\label{intro}
\end{figure}

Previous BFP quantization variants have varying degrees of optimization deployed to servers and edge devices. Nevertheless, previous researches on BFP-based quantization empirically rely on particular BFP representation and the block size, but such quantization settings are not necessarily the optimal solution in the quantization space, which offers limited opportunities for performance and efficiency gains. Specifically, Fig. \ref{intro} illustrates two selection strategies for BFP quantization parameters. Empirically-driven selection may result in suboptimal quantization outcomes, leading to inferior performance in terms of quantization loss and model size compared to search-based strategies. This is due to the large parameter search space in BFP quantization, where empirical configurations cannot robustly adapt to different DNNs. Therefore, searching for optimal BFP quantization parameters can effectively improve DNN efficiency.

To achieve high performance and energy efficiency, we provide a bitwidth-aware analytical searching framework (called ``BitQ'') to fully exploit the BFP quantization for resource-constrained devices. 
While ensuring the quantitative loss is acceptable,  we notice the data movement is a key bottleneck and dominates the energy consumption of DNN inference on resource-constrained devices \cite{ivanov2021data,chen2014diannao,gao2024dynamic}. For example, the memory access occupies $>90\%$ of the total energy for representative DNN deployments \cite{chen2014diannao}. Based on this observation, we formulate and resolve an optimization problem to identify the optimal BFP quantization configuration. Specifically, the model accuracy and the data movement volume can be utilized as the trade-off for the exploration of BFP quantization configurations. We develop a BFP-based modeling approach that can fully explore the data reuse of DNNs when evaluating the data movement volume. Experimental results based on image classification, object detection, and scene segmentation applications demonstrate that BitQ outperforms the state-of-the-art baselines.   

The contributions of this work are summarized as follows.

\begin{itemize}
\item We propose a BFP-based bitwidth-aware analytical searching framework (named "BitQ") tailored for optimizing DNN inference on embedded platforms by the trade-off between accuracy and data movement volume.

\item We involve formulating and solving an optimization problem to pinpoint the BFP block size and bitwidth distribution, carefully balancing accuracy with performance considerations by taking the block size into account, bitwidth configuration, and data reuse.

\item Experimental results demonstrate that our optimized bitwidth allocation approach for BFP DNNs outperforms state-of-the-art baselines, delivering better computational efficiency and decreased memory access demands, preserving the accuracy.
\end{itemize}

\section{Related Work}
In embedded platforms, there are significant advantages in terms of decreased latency and enhanced energy efficiency through various forms of model compression. 
 Pruning \cite{bai2023unified} and quantization \cite{kuzmin2024pruning} are two widely employed techniques in practice. Pruning techniques primarily focus on the removal of individual parameters from DNNs. While pruning is theoretically interesting, it poses greater implementation challenges within the hardware configurations \cite{pconv}. On the other hand, quantization decreases the bitwidth for weights and network calculations, resulting in both predictable memories saving and the reduction of necessary computation.

\begin{figure*}[!t]
\centering
\resizebox{\textwidth}{!}{
  \includegraphics{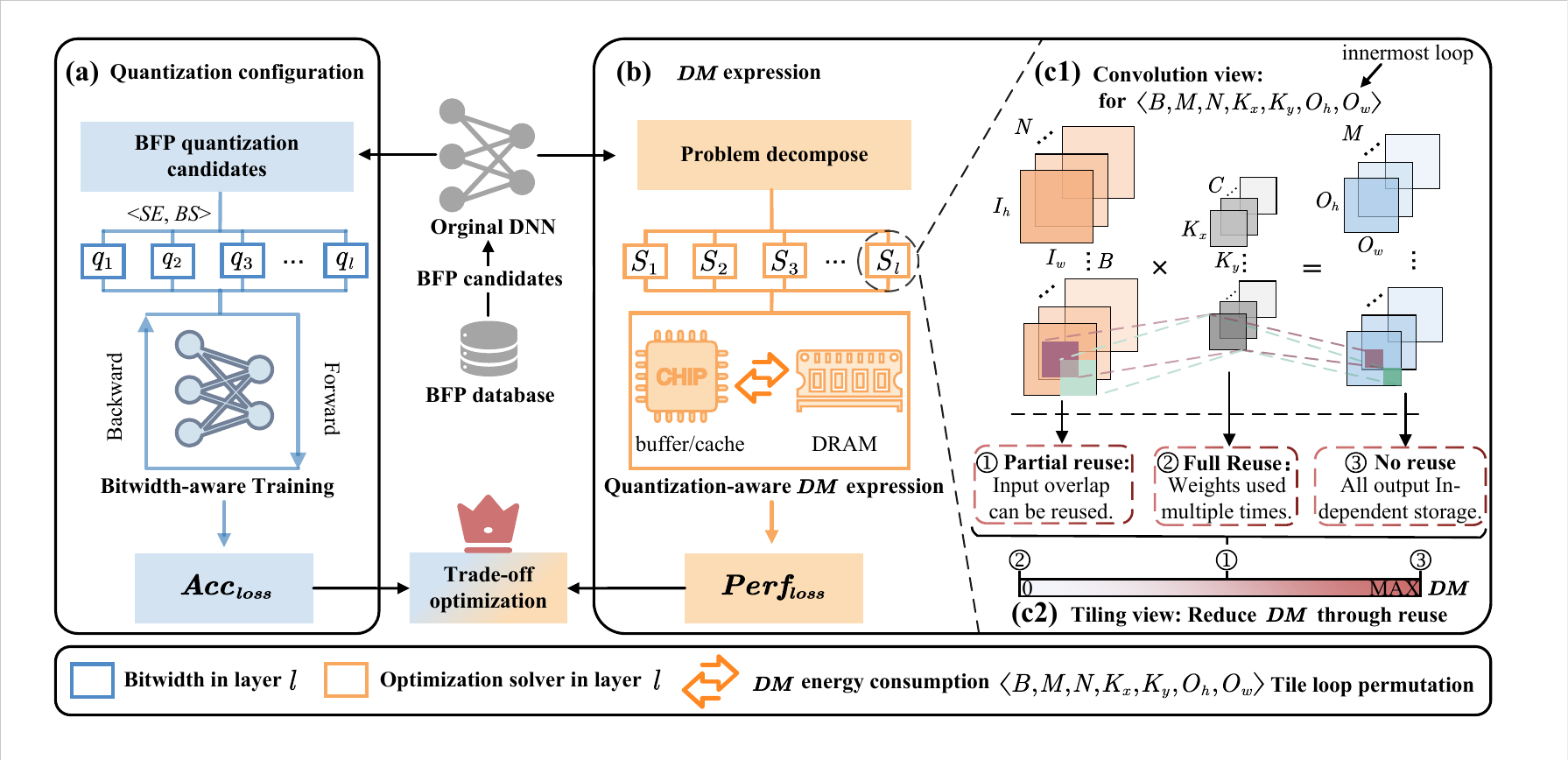}
}
\caption{Workflow of the trade-off optimization of BitQ. (a) BFP quantization configuration. (b) Data movement ($DM$) expression. And (c) basis of convolution (c1) and data reuse in tiling (c2).}
\label{mainframe}
\end{figure*}

The objective of DNN quantization is to quantize both weights and activations, enabling efficient calculation by leveraging limited hardware resources, which are faster and more memory-efficient compared with the standard FP32 format. There are two representative quantization formats: fixed-point and block floating point data representation \cite{yu2023boost,akkad2023embedded}. Fixed-point numbers are treated as integers in hardware, facilitating simple and fast operations, but incorporating a scaling offset enabling limited fractional precision. Differently, BFP quantization involves grouping several values with a shared exponent while preserving individual mantissa. Based on the shared exponent, BFP quantization can overcome the limited numerical range encountered by fixed-point arithmetic. Among various bitwidths of BFP formats, both 16-bit and 8-bit  quantization are widely deployed and utilized, owing to their efficiency in terms of memory storage, processing speed, and overall computational performance \cite{10472132,lian2019high}.

Recently, variant algorithms based on BFP quantization have effectively improved quantization performance. HBFP \cite{hbfp} is a hybrid approach combining BFP for dot products and floating point for other operations, offering high accuracy and superior hardware density. To provide adaptive BFP configurations for memory and calculation efficiency, DBPS \cite{lee2023dbps} and FlexBlock \cite{noh2023flexblock} incorporate dynamic block sizes and scalable BFP quantization precision to reduce energy consumption. FAST \cite{zhang2022fast} dynamically chooses the optimal BFP precision for both activations and weights during calculation. Additionally, BSFP \cite{bsfp} achieves high computation efficiency by quantizing each weight vector as a superposition of multiple subword vectors with scaling factors in low-bit floating-point (LBFP), effectively fitting the weight vector distribution.


\section{Methodology}
This section outlines the BitQ workflow and its 3 key components: BFP quantization configuration, bitwidth-aware data movement expression, and trade-off optimization.

\subsection{Workflow}

Fig. \ref{mainframe} depicts the workflow of BitQ, which consists of three stages: BFP quantization configuration, bitwidth-aware data movement ($DM$) expression, and trade-off optimization. The BFP quantization configuration stage provides a set of DNN model accuracy loss after different BFP quantization configurations. The bitwidth-aware $DM$ expression stage provides the data movement of convolutions. The trade-off optimization stage finds the best BFP quantization configuration based on the  minimization of the objective function ($\mathcal{O}$), which is formulated as:

\begin{equation}
\label{eq_object}
min:\ \mathcal{O} = Acc_{loss}(SE,BS) + \alpha \cdot Perf_{loss}(SE,BS,\vec{I},\vec{P})
\end{equation}

where $Acc_{loss}$ denotes the accuracy loss at the BFP quantization configuration stage. $Perf_{loss}$ corresponds to the performance loss related to data movement in the $DM$ expression stage. $\alpha$ serves as a balance factor for these losses. 
$SE$ is the bitwidth of the shared exponent. $BS$ reflects the block size of the shared exponent component.
Input activations, output activations, and filter weights each have their designated $SE$ and $BS$. $\vec{I}$ signifies the tiling size.
$\vec{P}=(\vec{p_1},...,\vec{p_l})$ denotes the loop permutations for tiling in the $l$-th convolution layer. 



\subsection{BFP Quantization Configuration}
Fig. \ref{mainframe}(a) depicts the BFP quantization configuration process, producing a range of accuracy candidates across different BFP quantization setups. The accuracy loss is generated by training. Each convolution layer within a DNN model is assigned a distinct bitwidth size that incorporates shared exponents and block sizes ($<\!\!SE, BS\!\!>$) throughout the training process. Input activations, output activations, and filter weights are individually tailored with their specific $<\!\!SE,BS\!\!>$ configurations.

\begin{figure}[!t]
\centering
\includegraphics{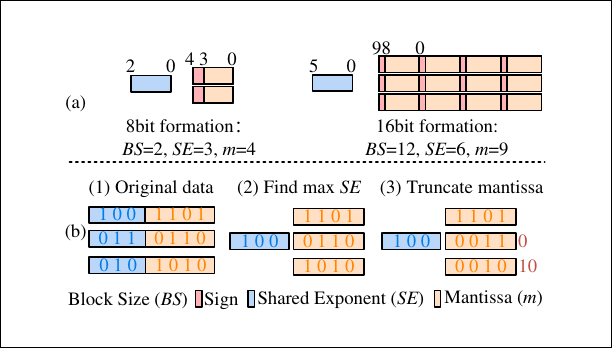}
\caption{BFP data representation. (a) Illustration of 8-bit and 16-bit BFP data representation. (b) Process of generating BFP data from original data.}
\label{bfp}
\end{figure}

Fig. \ref{bfp} depicts the fundamental organization of BFP data representation.  Fig. \ref{bfp}(a) illustrates the BFP data representation under 8-bit and 16-bit data widths, where each data block shares the same exponents. For example, in the 8-bit BFP data representation (with a block size of $BS=2$, shared exponent of $SE=3$, and mantissa $m=4$), two data points share a common exponent.   Fig. \ref{bfp}(b) illustrates the conversion process from original floating-point data to BFP representation, involving two key steps. Initially, it identifies the maximum exponent within a block size to serve as the shared exponent. Subsequently, it adjusts the mantissa through a shifting operation, refining the mantissa by subtracting individual mantissa values from the mantissa associated with the shared exponent.

\subsection{Bitwidth-aware Data Movement Expression}

Fig. \ref{mainframe}(b) depicts the stage of bitwidth-aware data movement ($DM$) expression, aiming to compute data movement for specific BFP configurations. The objective is to boost inference efficiency by reducing energy consumption. Since data movement plays a vital role in energy usage,  we employ $DM$ as a measure to evaluate inference efficacy. In particular, we establish the connection between performance decline ($Perf_{loss}$ from Equation (\ref{eq_object})) and $DM$, formulated as:

\begin{small}
\begin{equation}
\label{eq_pfloss}
Perf_{loss} = \frac{DM_{sum}}{DM_{max}}
\end{equation}
\end{small}
\begin{small}
\begin{equation}
\label{eq_dmsum}
DM_{sum} = \sum_{1 \leq l \leq L}{DM_{l}}
\end{equation}
\end{small}
where $DM_{sum}$ identifies the total volume of data movement across all convolution layers, as outlined in Equation (\ref{eq_dmsum}). $DM_l$ denotes the data movement within the $l$-th convolution layer ($1 \leq l \leq L$), where $L$ refers to the total number of convolution layers. $DM_{max}$ corresponds the maximum value attained by $DM_{sum}$ among various potential solutions encompassing different BFP quantization formats. The expression for the data movement in the $l$-th convolution layer ($DM_{l}$) is defined as,

\begin{small}
\begin{equation}
\label{eq_dm}
DM_{l}= \sum{DM_{i}}
\end{equation}
\end{small}
where $DM_{i}$ identifies the data movement volume in the innermost loop of Fig. \ref{mainframe}(c1),  calculated as,


\begin{small}
\begin{equation}
\label{eq_dminm}
DM_{i} = \left\{
\begin{aligned}
& It_{i} \times DF_{t}(q_i, q_o, q_w), && \text{no reuse} \\
& DF_{t}(q_i, q_o, q_w) + {} \\
& \quad (It_{i}-1) \times DF_{n}(q_i, q_o, q_w), && \text{partial reuse} \\
& 0, && \text{full reuse}
\end{aligned}
\right.
\end{equation}
\end{small}

where $DF_{t}$ is the data volumes of a single tile.  $DF_{n}$ signifies the effective volumes of new data exchanged between two consecutive tiles when data is partially reused. $It_{i}$ denotes the iteration count within the innermost loop of a convolution. $q_i$, $q_o$, and $q_w$ refer to the bitwidths of input activations, output activations, and filter weights. These can be computed as, 
\begin{small}
\begin{equation}
\label{eq_qw}
q_{x}=\left\{
\begin{aligned}
&(q_{b}-SE)+\frac{SE}{BS\cdot I_h\cdot I_w},\quad &&x=i\\
&(q_{b}-SE)+\frac{SE}{BS\cdot O_h\cdot O_w},\quad &&x=o\\
&(q_{b}-SE)+\frac{SE}{BS\cdot K_x\cdot K_y},\quad &&x=w
\end{aligned}
\right.
\end{equation}
\end{small}
where $q_b$ denotes the overall bitwidth of data (e.g., 16-bit or 8-bit ($q_{b}=16$ or $q_{b}=8$)). $q_x$ comprises a signed mantissa and a shared exponent, where the mantissa can be derived from $q_{b}-SE$, and the shared exponent is determined by the block size $BS$.




For a convolution layer, we account for data reuse effects in data movement. Fig.  \ref{mainframe}(c2) illustrates three forms of data reuse, denoted as $\text{\textcircled{1}}$,~$\text{\textcircled{2}}$, and $\text{\textcircled{3}}$. In the first scenario $\text{\textcircled{1}}$, there is no data reuse due to the lack of overlap in input activations when calculating adjacent output activations. Scenario $\text{\textcircled{2}}$ showcases partial input activation reuse resulting from convolution with a specific stride size. Scenario $\text{\textcircled{3}}$ displays complete data reuse where the kernel size is smaller than the input feature maps, allowing full reutilization for output activations. Consequently, $DM_{i}$ encompasses these three scenarios based on distinct data reuse patterns: no reuse during current computation, partial reuse due to convolution with a stride size, and complete data reuse eliminating the need for data movement. 



\subsection{Trade-off Optimization}


The trade-off optimization involves two key steps: identifying the optimal data movement as well as considering data reuse and striving for the most favorable solution for the final outcomes.

\textbf{Data Movement Optimization.} Within a specific memory setting, configuration of the quantization, and tiling loop permutation for the $l$-th convolution layer ($<\!\!SE, BS, \vec{p_l}\!\!>$), the optimization sub-problem aims to seek the best tiling size ${I_l}$. The minimum cost of $DM_l$ can be articulated as,

\begin{small}
\begin{equation}
\label{eq_sltiling}
min:\ DM_l\qquad s.t.\ \sum{DF_{t}(q_i, q_o, q_w)} \leq MC
\end{equation}
\end{small}
where $MC$ denotes the memory capacity constraint on embedded platforms. The calculation of $DM_l$ aligns 
 with Equation (\ref{eq_dm}). $DF_{t}$ encompasses the data footprint volume of three tiles (e.g., tiles of input activations, output activations, and filter weights), computed based on  Fig. \ref{mainframe}(c2).  


To minimize the total data movement cost, we optimize sub-problems within constraints. The optimization of each minimal single-level tiling problem, considering the constraint of $MC$, can be realized through nonlinear solvers \cite{wachter_2006_on}, 
which utilizes the feasible region to approximate the global optimum.

\textbf{Optimal Solution Search.} The trade-off optimization phase (the last stage of  Fig. \ref{mainframe}) focuses on seeking the best solution for the DNN inference while minimizing accuracy loss. The design space encompasses three parameter groups: BFP quantization strategies $(SE, BS)$; loop permutations $P$; and tiling sizes $I$. 

To identify the optimal solution, we have formulated an objective function that guides exploration within the design space, to balance model accuracy loss with inference performance. The objective function, as outlined in Equation (\ref{eq_object}), facilitates the simultaneous minimization of accuracy and performance losses. Additionally, a trade-off factor ($\alpha$) is employed to navigate the balance between model accuracy and inference efficiency. 

\section{Evaluations}

\subsection{Experiment Setup}
We conduct comprehensive experimental evaluations across diverse vision tasks, encompassing image classification on ImageNet-1K \cite{imagenet}, object detection and instance segmentation on COCO2017 \cite{COCO}, and semantic segmentation on ADE20K \cite{ADE}. Table \ref{t1} outlines the characteristics of the benchmarks and datasets.


Our approach includes two versions based on the size of $q_b$, 
namely $\text{BitQ}_{16}$ and $\text{BitQ}_{8}$  ($q_b=16$ and $q_b=8$, where $q_b$ is the  overall bitwidth of data), respectively. The balance factor $\alpha$ is set to be 0.2. The search process identifies the optimal configuration from Table \ref{t2} of BFP quantization candidates. We utilize a BFP quantizer for quantization-aware training, extending the training by 1,000 steps to compensate for the reduced data expression range due to quantization, following the recommended parameters. 

We provide the original 32-bit floating-point data representation (labeled as ”Original”) for comparison. Additionally, we compare our approach against representative 16-bit BFP-based baselines, including DBPS \cite{lee2023dbps} and Flexblock \cite{noh2023flexblock}, as well as 8-bit BFP-based baselines, FAST \cite{zhang2022fast} and BSFP \cite{bsfp}. The performance of the baselines comes from the results reported in the literature and our own reproductions.

The evaluation is conducted using a 2.90 GHz Intel i7-10700F CPU and an NVIDIA V100 graphics card. Models are quantified, trained, and validated on specific datasets over 10 rounds to derive the average performance metrics.

\begin{table}[!t]
\small  
\resizebox{0.48\textwidth}{!}{
\begin{tabular}{c|c|c}
\hline
Task              & Model                                   & Dataset                               \\ \hline
                  & GoogLeNet \cite{GoogLeNet}              &                                               \\ \cline{2-2}
Image             & ResNet-18 \cite{resnet18}               & ImageNet-1K      \\ \cline{2-2}
Classification    & ECA-MobileNetV2 \cite{ECA-MobileNetV2}  &  \cite{imagenet}                                             \\ \cline{2-2}
                  & ConvNeXt-L \cite{ConvNeXt-L}            &                                               \\ \hline
Object            & YOLOv8 \cite{YOLOv8}                    & COCO2017            \\ \cline{2-2}
Detection         & ONE-PEACE \cite{ONE-PEACE}              &  \cite{COCO}                                             \\ \hline
Instance          & YOLOv8 \cite{YOLOv8}                    & COCO2017            \\ \cline{2-2}
Segmentation      & MaskDINO \cite{MaskDINO}                & \cite{COCO}                                              \\ \hline
Semantic          & MaskDINO \cite{MaskDINO}                & ADE20K                       \\ \cline{2-2}
Segmentation      & ONE-PEACE \cite{ONE-PEACE}              & \cite{ADE}                                              \\ \hline
\end{tabular}%
}
\caption{Characteristic of benchmarks and datasets.}
\label{t1}
\end{table}

\begin{table}[!t]
\resizebox{0.48\textwidth}{!}{
\begin{tabular}{cc|c}
\hline
\multicolumn{2}{c|}{Shared Exponents ($SE$)}                              &Block Sizes ($BS$)                                              \\ \hline
\multicolumn{1}{c|}{$\text{BitQ}_{16}$}               & $\text{BitQ}_{8}$              & \multirow{2}{*}{\{1, 2, 4, 8, 16, 24, 32, 48\}} \\ \cline{1-2}
\multicolumn{1}{c|}{\{2, 3, 4, 5, 6, 7\}} & \{2, 3, 4, 5, 6\} &                                            \\ \hline
\end{tabular}
}
\caption{BFP quantization configuration candidates.}
\label{t2}
\end{table}


\begin{table*}[!t]

\small
\resizebox{1\textwidth}{!}{
\begin{tabular}{c|l|cl|l|ccc|ccc}
\toprule
 & \multicolumn{1}{c|}{} & \multicolumn{2}{c|}{} \vspace{-0.08cm}& \multicolumn{1}{c|}{32-bit} \vspace{0.01cm}& \multicolumn{3}{c|}{16-bit} \vspace{0.01cm}& \multicolumn{3}{c}{8-bit} \vspace{0.01cm}\\ \cline{5-11}\vspace{-0.645cm}\\ \\
\multirow{-2}{*}{Task} & \multicolumn{1}{c|}{\multirow{-2}{*}{Metric}} & \multicolumn{2}{c|}{\multirow{-2}{*}{Model}} & \multicolumn{1}{c|}{Original} & DBPS & FlexB & $\text{BitQ}_{16}$ & FAST & BSFP & $\text{BitQ}_{8}$ \\ \midrule
 &  & \multicolumn{2}{c|}{GoogLeNet} & 67.79\% & 66.71\% & 66.58\% & \textbf{67.49\%} & 66.31\% & 66.54\% & \textbf{66.58\%} \\
 &  & \multicolumn{2}{c|}{ResNet-18} & 69.70\% & 68.42\% & 68.29\% & \textbf{69.83\%} & 68.52\% & 69.67\% & \textbf{69.71\%} \\
 & \multicolumn{1}{c|}{Top-1 Acc.} & \multicolumn{2}{c|}{ECA-MobV2} & 72.56\% & 68.93\% & 70.25\% & \textbf{70.85\%} &\textbf{70.15\%} & 69.63\% & 69.85\% \\
 &  & \multicolumn{2}{c|}{ConvNeXt-L} & 85.82\% & 83.85\% & 83.87\% & \textbf{84.31\%} & 83.82\% & 84.18\% & \textbf{84.26\%} \\
\multirow{-5}{*}{\begin{tabular}[c]{@{}c@{}}Image \\ Classification\end{tabular}} &  & \multicolumn{2}{c|}{Gmean\_C} & 73.65\% & 71.67\% & 71.94\% & \textbf{72.84\%} & 71.89\% & 72.20\% & \textbf{72.29\%} \\ \midrule
 &  & \multicolumn{2}{c|}{YOLOv8} & 37.30\% & 36.71\% & 36.28\% & \textbf{36.78\%} & 36.36\% & 36.47\% & \textbf{36.50\%} \\
 & \multicolumn{1}{c|}{Box AP} & \multicolumn{2}{c|}{ONE-PEACE} & 60.40\% & 59.47\% & 59.34\% & \textbf{60.14\%} & 59.27\% & 59.23\% & \textbf{59.35\%} \\
\multirow{-3}{*}{\begin{tabular}[c]{@{}c@{}}Object \\ Detection\end{tabular}} &  & \multicolumn{2}{c|}{Gmean\_O} & 47.46\% & 46.72\% & 46.40\% & \textbf{47.03\%} & 46.42\% & 46.48\% & \textbf{46.54\%} \\ \midrule
 &  & \multicolumn{2}{c|}{YOLOv8} & 30.50\% & 30.03\% & 29.99\% & \textbf{30.19\%} & 29.76\% & \textbf{29.88\%} & 29.86\% \\
 & \multicolumn{1}{c|}{Mask AP} & \multicolumn{2}{c|}{MaskDINO} & 46.30\% & 46.28\% & 46.30\% & \textbf{46.31\%} & 45.72\% & 45.88\% & \textbf{45.99\%} \\
\multirow{-3}{*}{\begin{tabular}[c]{@{}c@{}}Instance \\ Segmentation\end{tabular}} &  & \multicolumn{2}{c|}{Gmean\_I} & 37.58\% & 37.28\% & 37.26\% & \textbf{37.39\%} & 36.89\% & 37.03\% & \textbf{37.06\%} \\ \midrule
 &  & \multicolumn{2}{c|}{MaskDINO} & 48.73\% & 48.28\% & 48.13\% & \textbf{48.37\%} & 48.26\% & 48.15\% & \textbf{48.34\%} \\
 & \multicolumn{1}{c|}{mIoU} & \multicolumn{2}{c|}{ONE-PEACE} & 62.27\% & 61.27\% & 61.36\% & \textbf{61.42\%} & 60.34\% & 60.37\% & \textbf{60.91\%} \\
\multirow{-3}{*}{\begin{tabular}[c]{@{}c@{}}Semantic \\ Segmentation\end{tabular}} &  & \multicolumn{2}{c|}{Gmean\_S} & 55.09\% & 54.39\% & 54.34\% & \textbf{54.51\%} & 53.96\% & 53.91\% & \textbf{54.26\%} \\ \bottomrule
\end{tabular}
}
\raggedright
\textbf{*}{ECA-MobV2 is denoted as ECA-MobileNetV2 \cite{ECA-MobileNetV2}, and FlexB as FlexBlock \cite{noh2023flexblock}.}
\caption{Validation accuracy metrics. Gmean\_C, Gmean\_O, Gmean\_I, and Gmean\_S identify the geometric mean of different models in image classification, object detection, instance segmentation, and semantic segmentation tasks, respectively. The bold data indicate the best value under 16-bit and 8-bit bitwidth settings.}

\label{i_acc}
\end{table*}



\subsection{Accuracy Comparison}
We conduct the accuracy evaluation involving representative four types of computer vision tasks. A geometric mean is supplied for accuracy evaluation within each task.

Table \ref{i_acc} lists the accuracy result. The proposed $\text{BitQ}_{16}$ and $\text{BitQ}_{8}$ show superior performance, winning all 14 and 12 out of 14 comparisons in the 16-bit and 8-bit bitwidth settings, respectively. 
Upon analyzing the baselines, we observe that DBPS \cite{lee2023dbps} and FlexBlock \cite{noh2023flexblock} accelerate the training convergence by utilizing dynamic block sizes. However, since the final block sizes are empirically determined, the training convergence does not necessarily translate to optimal inference performance. FAST \cite{zhang2022fast} employs Stochastic Rounding to implement dynamic training for Transformers, which leads to improved accuracy in some Transformer-based models. The criterion-optimal quantization flow of BSFP \cite{bsfp} explores the quantization parameter space only partially, as the exhaustive search would be computationally prohibitive.

When compared with the $\text{Original}$ baseline, $\text{BitQ}_{16}$ displays slight average accuracy decreases of 0.81\%, 0.43\%, 0.19\%, and 0.58\% across the same four benchmark groups. The slightly lower accuracy compared to the Original baseline is due to employing a smaller bitwidth for data representation.
Compared with $\text{BitQ}_{8}$, $\text{BitQ}_{16}$ on average exhibits slightly better accuracy performance with increases of  0.55\%, 0.49\%, 0.33\%, and 0.25\% across the four groups of benchmarks, owing to its larger bitwidth for data representation. Despite reducing the bitwidth from 16-bit to 8-bit BFP data representation, the accuracy loss remains 0.55\%, highlighting an efficient approach to reduce data movement.

\begin{figure}[!t]
\centering
\resizebox{0.5\textwidth}{!}{
  \includegraphics{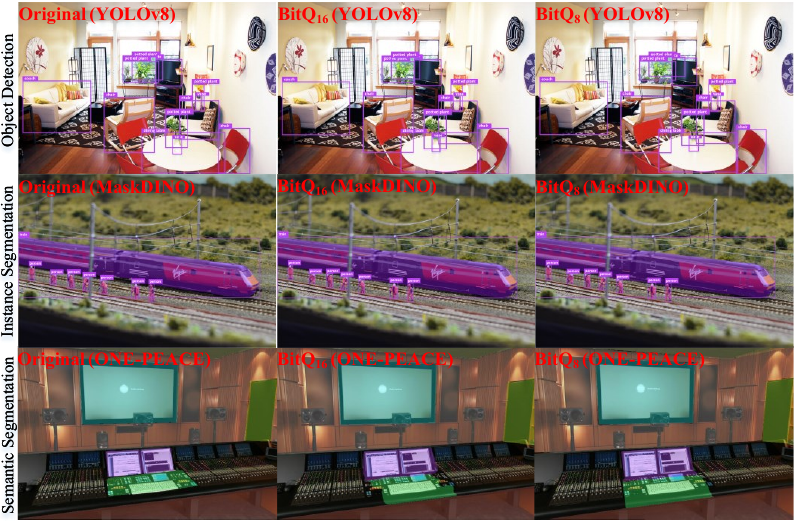}
}
\caption{Visualization results of Original and $\text{BitQ}_{16(8)}$ on downstream tasks. 
} 
\label{vis}
\end{figure}

Fig. \ref{vis} illustrates the visualization outcomes, contrasting $\text{BitQ}_{16(8)}$ with the Original. 
We employ the images in the COCO2017 \cite{COCO} for  Object Detection and Instance Segmentation tasks and the image in the ADE20K \cite{ADE} dataset is utilized for Semantic Segmentation tasks. There is no significant difference by comparing the results produced by the original model and the original model applying the $\text{BitQ}_{16(8)}$ quantization technique. 

\begin{figure*}[!t]
\centering
\resizebox{1\textwidth}{!}{
  \includegraphics{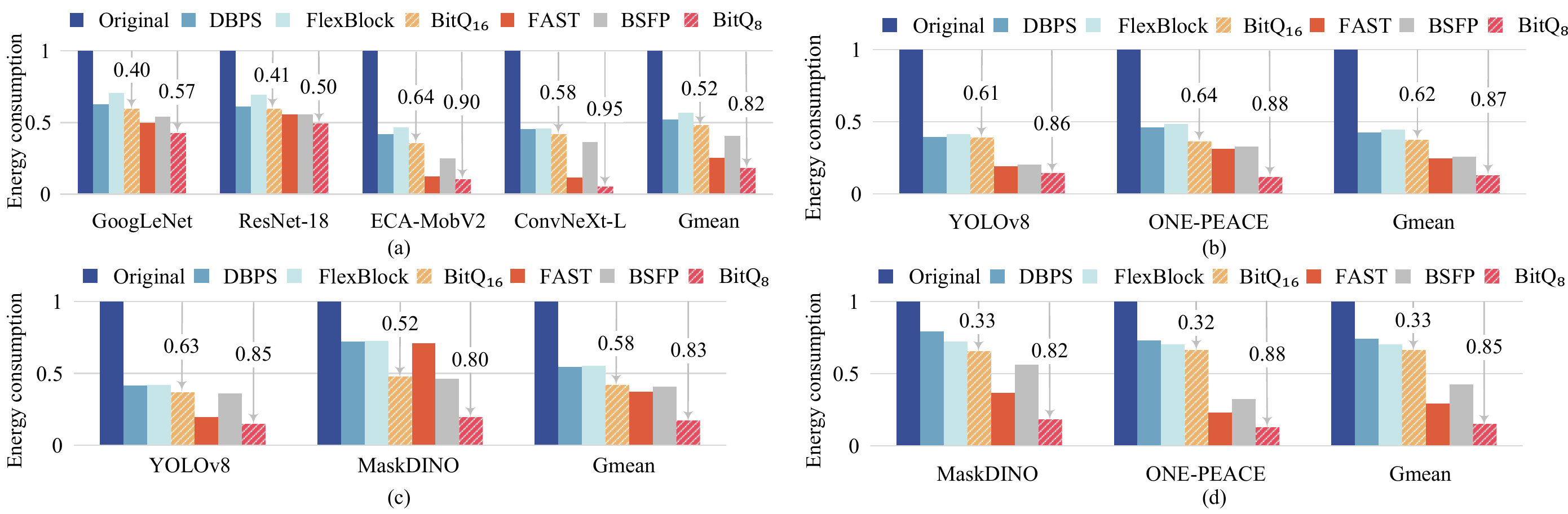}
}
\caption{Normalized energy comparison under (a) image classification, (b) object detection, (c) instance segmentation, and (d) semantic segmentation. Less energy consumption is preferable. Gmean identifies the geometric mean across various models for corresponding visual tasks.} 
\label{eng}
\end{figure*}


\begin{figure*}[!tb]
\centering
\resizebox{1.0\textwidth}{!}{
  \includegraphics{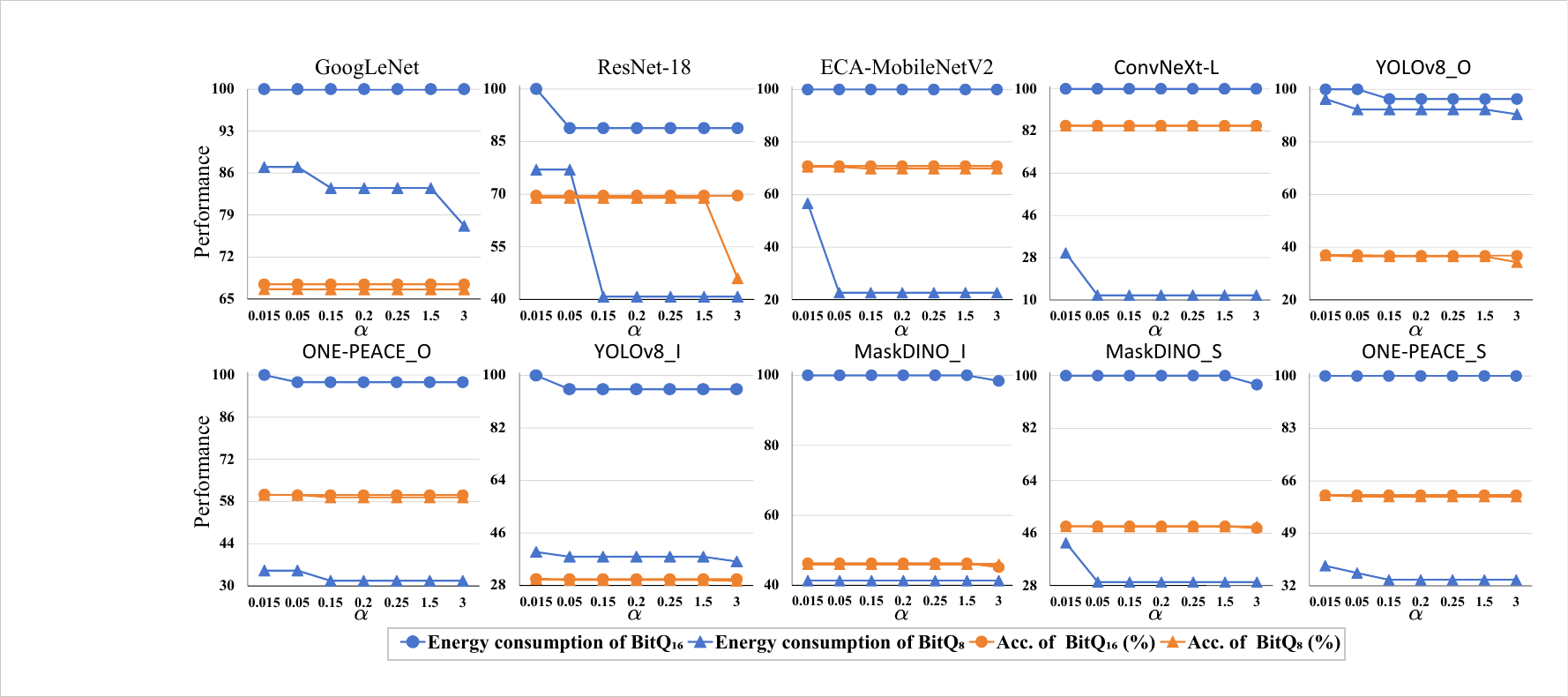}
}
\caption{Accuracy and energy consumption sensitivity of $\text{BitQ}_{16(8)}$ with varying values of $\alpha$ . The suffixes \_O, \_I, and \_S indicate the model's application scenarios in object detection, instance segmentation, and semantic segmentation, respectively. For a better display, we normalize the energy consumption values to a maximum of 100.} 
\label{S_16+8}
\end{figure*}

\subsection{Energy Comparison}
Fig. \ref{eng} shows the normalized energy consumption under four groups of benchmarks. The energy consumption is calculated based on the volume of data movement:
\begin{equation}
\label{energy}
Energy = DM_S\times E_S + DM_D\times E_D
\end{equation}
where $DM_S$ and $DM_D$ indicate the quantities of data movement in static random access memory (SRAM) and dynamic random access memory (DRAM), respectively. 
$E_S$ and $E_D$ denote the energy consumption per unit of data movement in SRAM and DRAM, respectively. $E_S=0.16 \text{ pJ/bit}$ and $E_D=20 \text{ pJ/bit}$ \cite{horowitz20141}.
To provide a visual comparison, all energy consumption values are normalized to the $\text{Original}$ baseline, where the energy consumption is set to 1. 

When compared to Original, 
$\text{BitQ}_{16}$  averages energy consumption of 0.52$\times$, 0.62$\times$, 0.58$\times$, and 0.33$\times$ for the four benchmark groups. And $\text{BitQ}_{8}$ uses 0.82$\times$, 0.87$\times$, 0.83$\times$, and 0.85$\times$ of the energy across the same four benchmark groups when compared with the $\text{Original}$ baseline. 
The better energy saving of $\text{BitQ}_{16}$ and $\text{BitQ}_{8}$ compared to the baseline with the same bitwidth setting results from their effective quantization exploration for energy saving, as the shared exponent does not require reloading after its initial loading, and the data for inference calculation is fully reused.
Compared with $\text{BitQ}_{16}$, $\text{BitQ}_{8}$ on average consumes 0.30$\times$, 0.25$\times$, 0.25$\times$, and 0.52$\times$ for the four groups of benchmarks. The reduced energy consumption comes from the reduction in data bitwidth from 16-bit to 8-bit data representation.

\subsection{Sensitivity Analysis}

Within the BitQ framework, a pivotal parameter is the loss balance factor $\alpha$. Varied values of this parameter have the potential to impact inference performance and energy consumption diversely. Specifically, $\alpha$ plays a key role in the selection of BFP quantization parameters by striking a balance between accuracy and performance losses. We conduct the sensitivity analysis under different levels of $\alpha$, by seven representative values ($\alpha \in$~\{~0.015, 0.05, 0.15, 0.2, 0.25, 1.5, 3~\}, ranging from 0 to 3). We conduct the sensitivity analysis based on the same ten benchmarks (listed in Table \ref{t1}) for both 16-bit and 8-bit BFP data representation.

Fig. \ref{S_16+8} depicts the accuracy and energy consumption sensitivity of $\text{BitQ}_{16(8)}$ across different levels of $\alpha$. Energy consumption is normalized with respect to $\alpha=0.015$, To facilitate comparison, we scale the normalized energy consumption from 0 to 100, aligning both accuracy and normalized energy consumption on a common scale for clarity. 

It can be observed from Fig. \ref{S_16+8} that: 1) when $\alpha$ ranges from 0.05 to 3, BitQ achieves identical energy saving; 2) BitQ attains equivalent accuracy performance when $\alpha$ ranges from 0.015 to 3.
The experimental results also reflect that BitQ with 8-bit data representation achieves equivalent accuracy and energy advantages within a narrower $\alpha$ range (0.15–1.5 for 8-bit data representation compared to 0.05–3 for 16-bit data representation). After thorough analysis, considering the impact of various DNNs and datasets, we choose $\alpha$ as 0.2 as a balanced approach.

In summary, BitQ stands out as a better choice owing to its adept balance between energy saving and accuracy for DNN inference on resource-constrained devices.


\begin{table}[!t]
\resizebox{0.48\textwidth}{!}{
\begin{tabular}{c|cc|cc}
\toprule
\multirow{2}{*}{Method}     & \multicolumn{2}{c|}{16-bit}             & \multicolumn{2}{c}{8-bit}             \\ \cmidrule(lr){2-3} \cmidrule(lr){4-5} 
                            & \multicolumn{1}{c}{Acc.}    & Eng.   & \multicolumn{1}{c}{Acc.}    & Eng.   \\ \midrule
BitQ                    & \multicolumn{1}{c}{69.83\%} & 0.143J & \multicolumn{1}{c}{69.71\%} & 0.119J \\ 
w/o QAT                     & \multicolumn{1}{c}{66.82\%} & 0.121J & \multicolumn{1}{c}{31.74\%} & 0.066J \\ 
w/o $DM$ Expression         & \multicolumn{1}{c}{69.84\%} & 0.197J & \multicolumn{1}{c}{69.73\%} & 0.142J \\ 
w/o Trade-off               & \multicolumn{1}{c}{69.43\%} & 0.175J & \multicolumn{1}{c}{68.84\%} & 0.136J \\ \bottomrule
\end{tabular}
}
\raggedright
\textbf{*}{Eng. is denoted as energy consumption.}
\caption{Ablation results of BitQ on ResNet-18. }
\label{t4}
\end{table}

\subsection{Ablation Study}
To further validate the effectiveness of BitQ, we conduct ablation analysis. Since BitQ consists of three components, we perform ablations on each module individually. The specific results are shown in Table~\ref{t4}. In the w/o QAT method, we omit quantization-aware training, selecting quantization parameters solely based on $DM$ size. This results in suboptimal accuracy but minimizes energy consumption. Conversely, in the w/o $DM$ Expression method, we exclude data movement simulation during inference, leading to quantization parameters chosen solely for accuracy. While this method achieves optimal accuracy, it increases energy consumption. In the w/o Trade-off method, we independently consider accuracy loss and performance loss, using a Pareto optimal solution for parameter configuration. This method often fails to balance the two, resulting in suboptimal outcomes. Across all ablation settings, the BitQ algorithm consistently yields superior results.


\section{Conclusion}
We present a block floating point (BFP) based bitwidth-aware analytical modeling framework for optimizing DNN inference on resource-constrained devices. We establish and address an optimization challenge to determine the best BFP quantization configurations. Experimental results demonstrate that BitQ can achieve better energy saving, preserving accuracy loss on representative benchmarks.



\bibliography{aaai25}

\end{document}